\documentclass[english,runningheads]{llncs}
\usepackage[T1]{fontenc}
\usepackage[utf8]{inputenc}
\usepackage{float}
\usepackage{booktabs}
\usepackage{textcomp}
\usepackage{url}
\usepackage{amstext}
\usepackage{graphicx}
\usepackage{microtype}

\makeatletter

\providecommand{\tabularnewline}{\\}

\@ifundefined{showcaptionsetup}{}{%
 \PassOptionsToPackage{caption=false}{subfig}}
\usepackage{subfig}
\makeatother

\usepackage{babel}
\begin{document}
\title{Acute Lymphoblastic Leukemia Classification from Microscopic Images
using Convolutional Neural Networks}
\titlerunning{ALL Classification from Microscopic Images using CNNs}
\author{Jonas Prellberg \and Oliver Kramer}
\authorrunning{J. Prellberg and O. Kramer}
\institute{University of Oldenburg, Oldenburg, Germany\\
\email{\textnormal{\{}jonas.prellberg,oliver.kramer\textnormal{\}}@uni-oldenburg.de}}
\maketitle
\begin{abstract}
Examining blood microscopic images for leukemia is necessary when
expensive equipment for flow cytometry is unavailable. Automated systems
can ease the burden on medical experts for performing this examination
and may be especially helpful to quickly screen a large number of
patients. We present a simple, yet effective classification approach
using a ResNeXt convolutional neural network with Squeeze-and-Excitation
modules. The approach was evaluated in the C-NMC online challenge
and achieves a weighted F1-score of 88.91\,\% on the test set. Code
is available at\linebreak{}
\url{https://github.com/jprellberg/isbi2019cancer}.
\end{abstract}

\section{Introduction}

Acute lymphoblastic leukemia (ALL) is a blood cancer that is characterized
by the proliferation of abnormal lymphoblast cells, eventually leading
to the accumulation of a lethal number of leukemia cells \cite{Pui2017}.
If ALL is diagnosed in an early stage, treatment is possible. Diagnosis
is typically performed using a complete blood count and morphological
analysis of cells under a microscope by a medical expert. Flow cytometry
can replace this manual work but requires expensive equipment which
is not available everywhere. Therefore, automated systems that can
perform diagnosis using comparatively low-cost microscopic images
provide a great advantage.

To further research in this direction, public datasets are necessary
to compare different approaches and track the state-of-the-art. A
popular example for single-cell ALL classification is ALL-IDB2 \cite{ALL-IDB2},
but with only 260 images of white blood cells the dataset is too small
to properly take advantage of recent deep learning approaches. In
2018, a new dataset with more than 10,000 training images and a separate
test set of normal B-lymphoid precursors and malignant B-lymphoblasts
has been released as an online challenge\footnote{\url{https://competitions.codalab.org/competitions/20395}}
open to the public. The large size of this new dataset allows to create
improved classifiers based on deep neural networks and also provides
a more reliable comparison of competing approaches.

In this work we present our approach to the classification of healthy
and malignant cells on the mentioned dataset using a convolutional
neural network. In Section~\ref{sec:Dataset} the dataset is described
in more detail and our data augmentation strategy is outlined. The
convolutional neural network, a ResNeXt-variant with Squeeze-and-Excitation
modules, is presented in Section~\ref{sec:Network-architecture}.
In Section~\ref{sec:Experiments} we describe the training process
and show experimental results. A summary of related work and a conclusion
follow.

\section{Dataset\label{sec:Dataset}}

The challenge dataset \cite{dataset-1,dataset-2,dataset-3,dataset-4,dataset-5},
hereafter referred to as C-NMC dataset, contains images of white blood
cells taken from 154 individual subjects, 84 of which exhibit ALL.
Table~\ref{tab:Composition-of-dataset} provides a detailed breakdown
of the number of subjects and cells in training and test sets. The
dataset is imbalanced with about twice as many ALL cells as normal
cells.

\begin{figure}
\subfloat[]{\includegraphics[width=0.24\textwidth]{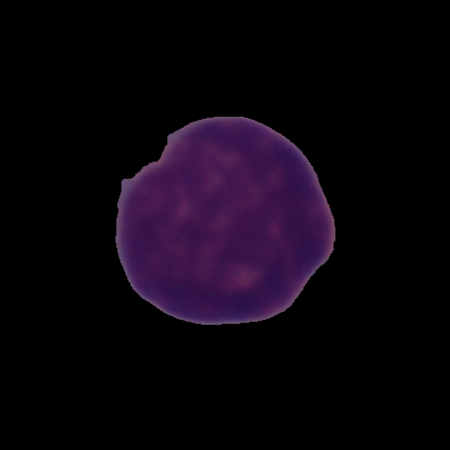}

}\subfloat[]{\includegraphics[width=0.24\textwidth]{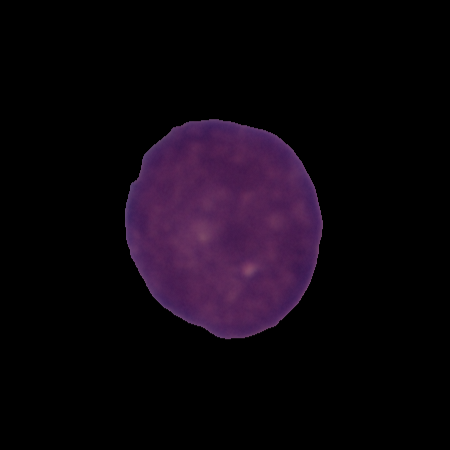}

}\subfloat[]{\includegraphics[width=0.24\textwidth]{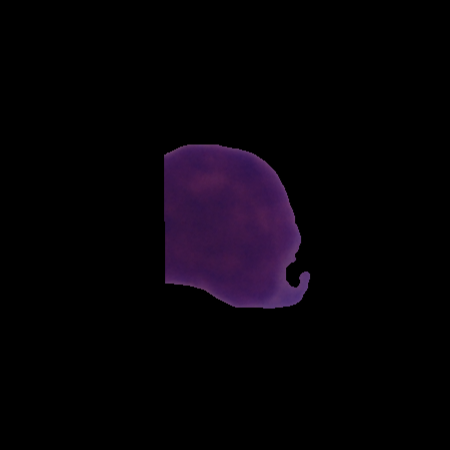}

}\subfloat[]{\includegraphics[width=0.24\textwidth]{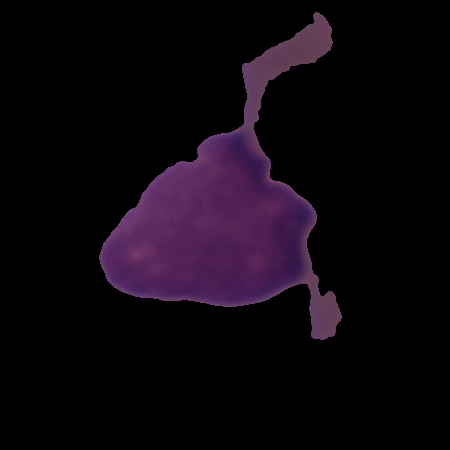}

}
\centering{}\caption{\label{fig:Example-images}Example images taken from the training
set. \textbf{(a)}~ALL cell \textbf{(b)}~Normal cell \textbf{(c)}~ALL
cell with part of the cell cut off due to an imperfect segmentation
\textbf{(d)}~Normal cell with superfluous background due to an imperfect
segmentation}
\end{figure}

Each image has a resolution of $450\times450$ pixels and contains
only a single cell as a consequence of preprocessing steps applied
by the dataset authors: An automated segmentation algorithm has been
used to separate the cells from the background. Each pixel that was
determined not to be part of the cell is colored completely black.
However, since the segmentation algorithm is not perfect, there are
instances where parts of the cell are inadvertently colored black
or superfluous background is included. Additionally, all images have
been preprocessed with a stain-normalization procedure that performs
white-balancing and fixes errors introduced due to variations in the
staining chemical \cite{dataset-2}. See Figure~\ref{fig:Example-images}
for example images from the dataset.

\noindent 
\begin{table}
\caption{\label{tab:Composition-of-dataset}Composition of the dataset. The
test set is unreleased and can only be evaluated against using an
online service so its composition is unknown.}

\centering{}\vspace{2mm}
\setlength{\tabcolsep}{5pt}%
\begin{tabular}{lcccc}
\toprule 
Dataset part & ALL subjects & Normal subjects & ALL cells & Normal cells\tabularnewline
\midrule
\midrule 
Train & 47 & 26 & 7272 & 3389\tabularnewline
\midrule 
Prelim. test & 13 & 15 & 1219 & 648\tabularnewline
\midrule 
Final test & 9 & 8 & ? & ?\tabularnewline
\bottomrule
\end{tabular}
\end{table}

\noindent 

Even though the dataset contains more than 10,000 images, several
data augmentation techniques can be applied to increase the amount
of training data further and improve the training of our convolutional
neural network. Since microscopic images are invariant to flips and
rotations, we perform horizontal and vertical flips with 50\,\% probability
each and rotations with an angle from $\left[0,360\right)$ degrees
chosen uniformly at random. Since convolutional neural networks with
pooling operations or strides larger than one are not perfectly translation
invariant, we also perform random translations of up to 20\,\% of
each side-length in horizontal and vertical directions. 

We do not randomly scale the images because cell size may be a diagnostic
factor to differentiate between ALL and normal cells \cite{Chiaretti2014}.
Furthermore, we do not apply any brightness or color augmentation
due to this dataset's stain-normalization preprocessing. Both data
augmentation methods are commonly used but would lead to an unnecessary
distribution shift between training and test set on this specific
dataset.

Additionally, the images are center-cropped to $300\times300$ pixels
to decrease the dimensionality of the input data. This will generally
make learning a classifier faster and easier. Even though the cropping
discards large parts of the image, it has no effect on the classification
accuracy because only very few cells are actually larger than this
crop. In many cases, images that are not completely black outside
of the crop are segmentation failures that include parts of the background.

\section{Network Architecture\label{sec:Network-architecture}}

Image classification benchmarks have driven the creation of many powerful
convolutional neural network architectures. We pick one of the recent
top-performing networks, a ResNeXt50 \cite{Xie_2017_CVPR}, as our
base model. The network consists of five convolutional stages with
spatial downsampling by a factor of 2 in between stages, followed
by global average pooling and a linear classifier. Each stage is made
from stacked building blocks, each of which computes a function of
the form 
\[
x'=x+\sum_{i=1}^{C}\mathcal{T}_{i}\left(x\right)
\]
 where $C$ is called the cardinality that controls the number of
parallel paths in a block and $\mathcal{T}_{i}\left(x\right)$ is
a function that projects $x$ into a lower-dimensional space, transforms
it and projects back into a space of the original dimensionality.
The functions $\mathcal{T}_{i}\left(x\right)$ are implemented using
a sequence of $1\times1$, $3\times3$ and $1\times1$ convolutional
layers. The block function is modified to 
\[
x'=x+\mathsf{SE}\left(\sum_{i=1}^{C}\mathcal{T}_{i}\left(x\right)\right)
\]
 where $\mathsf{SE}$ is the Squeeze-and-Excitation operation introduced
in \cite{Hu_2018_CVPR}, which has been shown to improve performance
on image classification benchmarks. In general, the $\mathsf{SE}$
operation learns a channel-wise rescaling of feature maps. First,
global-average-pooling is applied to the input feature map in order
to spatially aggregate the information of each channel. The resulting
vector is fed into a two-layer fully-connected network that outputs
scale factors. These factors are then used to scale each channel of
the original input feature map. More details on the SE-ResNeXt50 architecture
can be found in \cite{Hu_2018_CVPR}.

\section{Experiments\label{sec:Experiments}}

The previously described network architecture is used to classify
the C-NMC dataset. Since the dataset is imbalanced the overall accuracy
can be misleading and we report accuracy, sensitivity and specificity
as well as weighted F1-scores, weighted precision and weighted recall.
For sensitivity and specificity the ALL-class is regarded as the positive
class. Furthermore, because the data is from multiple subjects, we
report subject-level accuracies.

Hyperparameters for the training procedure have been chosen by validating
on the preliminary test set. The chosen hyperparametes are used for
multiple training runs with different random seeds and the best model
according to the F1-score on the preliminary test set is selected
for our competition entry. Final results on the test set are taken
from the online challenge leaderboard.

\subsection{Training and Testing}

The network is pre-trained on ImageNet \cite{ILSVRC15} and then fine-tuned
on the C-NMC training set. The loss function is the weighted binary
cross-entropy 
\[
l\left(y,\hat{y}\right)=-wy\ln\left(\sigma\left(\hat{y}\right)\right)-\left(1-y\right)\ln\left(1-\sigma\left(\hat{y}\right)\right)
\]
where $\sigma$ is the sigmoid function, $y$ is the true label, $\hat{y}$
is the network output and $w=n_{n}/n_{p}$ is the ratio of negative
to positive examples in the training set. The weight $w$ helps to
deal with the class imbalance present in the dataset.

The network is fine-tuned for 6 epochs using Adam \cite{adam} with
a batch size of 16. The learning rate is decayed using a step function
that starts at $\eta_{\text{base}}=1$ and is divided by 10 every
2 epochs. This learning rate is multiplied with factors specific to
the part of the network that it is applied to: The first and second
stages\footnote{Stage refers to a stack of building blocks that operate on the same
spatial resolution as described in Section~\ref{sec:Network-architecture}
and \cite{Xie_2017_CVPR}.} have an effective learning rate of $\eta_{12}=10^{-6}\eta_{\text{base}}$,
the third to fifth stages use $\eta_{345}=10^{-4}\eta_{\text{base}}$
and the last fully connected layer uses $\eta_{\text{fc}}=10^{-2}\eta_{\text{base}}$.

This results in lower effective learning rates in the earlier layers
of the network and is desirable because the network is initialized
with weights from ImageNet pre-training. These have been shown to
contain generally useful image filters in the lowest layers \cite{Yosinski:2014:TFD:2969033.2969197}
and continuing to optimize these weights on a small dataset with large
learning rates will degrade these general filters.

During inference, we present 8 rotated versions of each image and
average the network output to further improve classification results.

\subsection{Model Selection}

We find that results are sensitive to the random seed, despite all
networks starting from the same weight initialization due to the pre-training.
Therefore, we conduct 24 training runs with different random seeds
and measure the F1-score on the preliminary test set over the course
of the training. We collect the 24 model checkpoints that have the
highest F1-scores in their respective training run and report results
in Table~\ref{tab:Results-proposal}.

\begin{figure}[H]
\vspace{-0.25cm}

\begin{centering}
\includegraphics[width=1\textwidth]{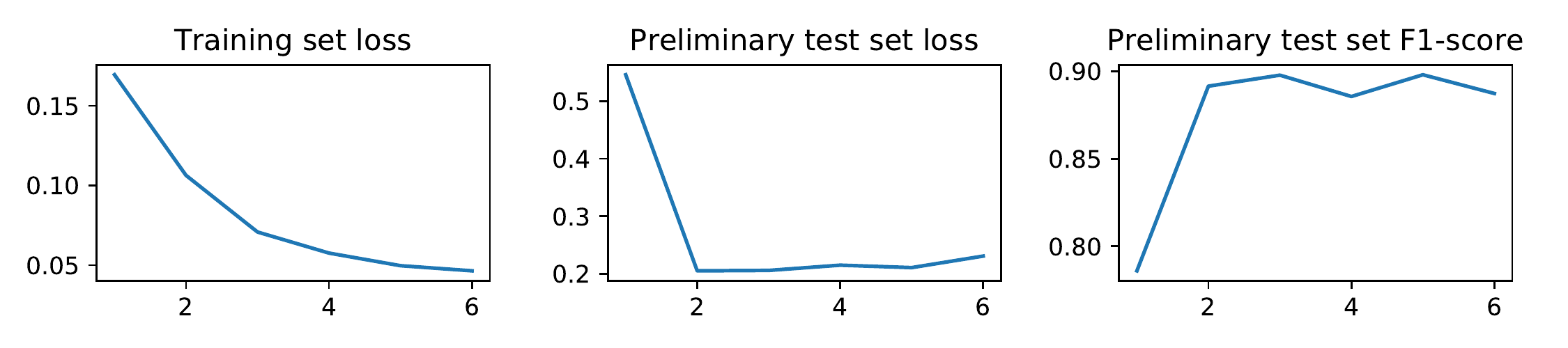}
\par\end{centering}
\caption{\label{fig:Train-curves}Training and validation curves of the best
model after each training epoch. The model achieves the maximum F1-score
after the 5th training epoch.}

\vspace{-1.25cm}
\end{figure}

\begin{figure}
\begin{centering}
\includegraphics[width=1\textwidth]{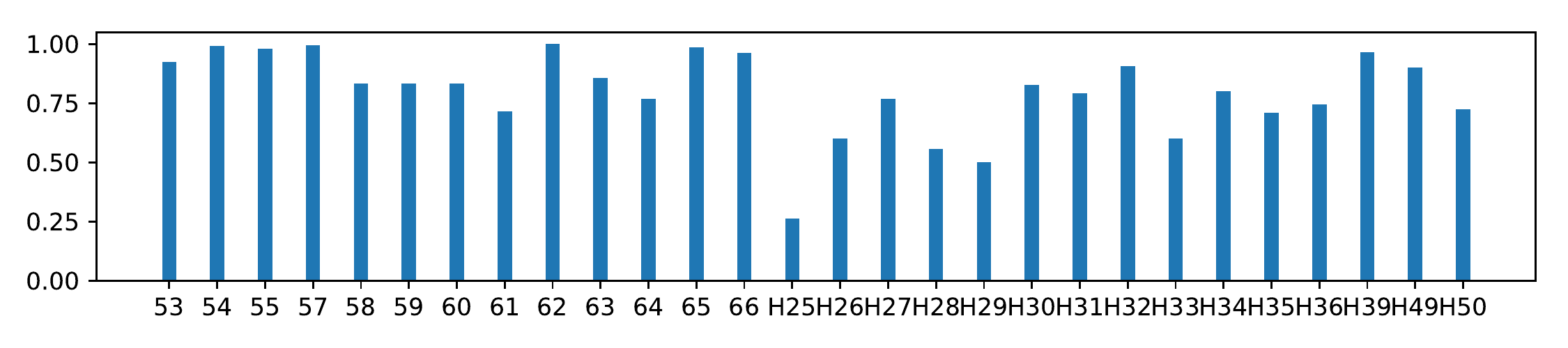}
\par\end{centering}
\caption{\label{fig:Subject-level-accuracy}Subject-level cell classification
accuracy on preliminary test set using the best model. Subjects with
the prefix H are healthy.}
\end{figure}

The best model achieves an F1-score of 89.81\,\% on the preliminary
test set. Figure~\ref{fig:Train-curves} shows the evolution of loss
and F1-score during its training process, while Figure~\ref{fig:Subject-level-accuracy}
shows cell classification accuracy grouped by subject. Overall, classification
for healthy subjects is worse because the specificity is comparatively
low. Consequently, false-positives can be expected when using the
network for diagnosis of patients. Still, false-positives are preferable
to false-negatives when trying to diagnose cancer with an automated
method.

\noindent 

\noindent 

\pagebreak{}

\noindent 
\begin{table}
\caption{\label{tab:Results-proposal}Results on the preliminary test set over
24 training runs using the best model of each run measured by F1-score}

\centering{}\vspace{2mm}
\setlength{\tabcolsep}{5pt}%
\begin{tabular}{llllllll}
\toprule 
 & Min. & Mean\,\textpm \,Std. & Max. &  & Min. & Mean\,\textpm \,Std. & Max.\tabularnewline
\midrule
\midrule 
Accuracy & 86.40 & 87.96\,\textpm \,0.90 & 89.88 & F1-score & 86.28 & 87.89\,\textpm \,0.90 & 89.81\tabularnewline
\midrule 
Sensitivity & 88.43 & 92.01\,\textpm \,1.44 & 94.50 & Precision & 86.27 & 87.91\,\textpm \,0.90 & 89.81\tabularnewline
\midrule 
Specificity & 75.31 & 80.36\,\textpm \,2.39 & 84.72 & Recall & 86.40 & 87.96\,\textpm \,0.90 & 89.88\tabularnewline
\bottomrule
\end{tabular}
\end{table}

\subsection{Results on the Final Test Set}

We use the selected model to classify the final test set and submit
the result to the online evaluation service. The model achieves a
weighted F1-score of 88.91\,\%.

\subsection{Ablation Studies}

\begin{figure}
\begin{centering}
\includegraphics[width=1\textwidth]{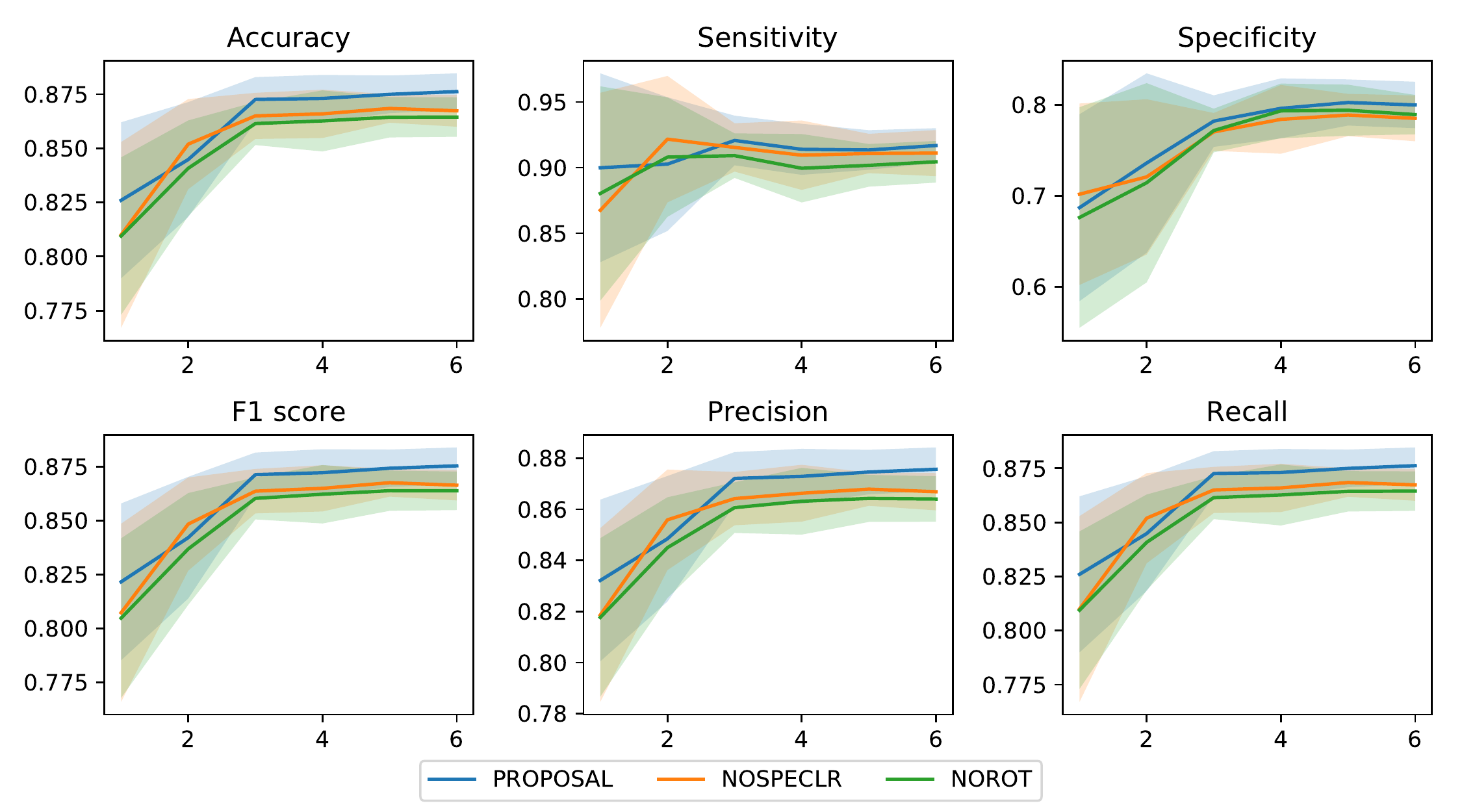}
\par\end{centering}
\caption{\label{fig:Accuracy-results}Metrics on the preliminary test set measured
after each training epoch for the proposed setting and the two ablation
studies. Lines display the mean over 24 training runs and the shaded
area marks one standard deviation. Best viewed electronically and
zoomed in.}
\end{figure}

The main design choices we made for the training and testing procedures
are layer-specific learning rates and test-time rotations. In order
to show their impact we perform two ablation studies:
\begin{description}
\item [{NOROT}] The training procedure is unchanged but during testing
the images are only presented in their original orientation. 
\item [{NOSPECLR}] During training, every layer has an effective learning
rate of $\eta_{\text{all}}\eta_{\text{base}}$ while still using the
scheduled decay for $\eta_{\text{base}}$ as described previously.
We test $\eta_{\text{all}}\in\left\{ 10^{-3},10^{-4},10^{-5}\right\} $
and report results for the best setting $\eta_{\text{all}}=10^{-4}$.
The testing procedure is unchanged.
\end{description}
Figure~\ref{fig:Accuracy-results} shows results on the preliminary
test set for our proposed setting (PROPOSAL) and the two ablation
studies. As expected, both ablations decrease performance in terms
of all considered metrics. Not using test-time rotations significantly
($p<0.001$, one-sided Mann-Whitney U test) decreases the mean F1-score
from 87.89\,\textpm \,0.90\,\% to 86.92\,\textpm \,0.81\,\%.
The influence of layer-wise learning rates is significant ($p<0.002$)
but smaller: A constant learning rate for all layers decreases the
mean F1-score to 87.20\,\textpm \,0.70\,\%.

\noindent 

\section{Related Work}

Previous work on automated ALL diagnosis from images can roughly be
divided into more recent approaches that use convolutional neural
networks as feature extractors and older approaches that use handcrafted
features.

\textbf{CNN features.} Rehman et al. \cite{Rehman2018} classify ALL
subtypes on a private dataset of 330 images using a pre-trained AlexNet
and fine-tuning. Shafique and Tehsin \cite{Shafique2018} classify
ALL subtypes on ALL-IDB augmented with 50 private images, also using
a pre-trained AlexNet and fine-tuning. Vogado et al. \cite{Vogado2017}
classify ALL on ALL-IDB using a number of different pre-trained CNNs
as fixed feature extractors. From these CNN features the most informative
ones are selected using PCA and finally classification is performed
with an ensemble of SVM, MLP, and random forest.

\textbf{Handcrafted features.} Putzu and Ruberto \cite{Putzu2013}
use ALL-IDB and classify a number of hand-crafted features like area,
compactness, roundness or area ratio between cytoplasm and nucleus
with an SVM. Mohapatra et al. \cite{Mohapatra2014} and Madhloom et
al. \cite{Madhloom2012} both use (different) private datasets and
classify using an ensemble of naive Bayes, KNN, MLP, SVM and a KNN
classifier respectively.

All works report good results but it is hardly possible to compare
them because the private datasets are unavailable and even on the
public ALL-IDB dataset researchers employ their own evaluation procedures.
Furthermore, in all cases the employed datasets are small, containing
a couple hundred images at most. It is important to have a large dataset
that can be used to track the state-of-the-art and we hope that this
role can be filled by the C-NMC dataset.

\section{Conclusion}

We present a simple, yet effective method to automatically classify
white blood cell microscopic images into normal B-lymphoid precursors
and malignant B-lymphoblasts. A recent convolutional neural network
architecture which is fine-tuned on the C-NMC dataset achieves promising
classification performance. We hope that the dataset authors release
a version with raw images that have not been preprocessed by their
in-house methods to allow further research that is not specific to
this preprocessing pipeline.

\bibliographystyle{splncs04}
\bibliography{isbi2019}

\begin{thebibliography}{10}
\providecommand{\url}[1]{\texttt{#1}}
\providecommand{\urlprefix}{URL }
\providecommand{\doi}[1]{https://doi.org/#1}

\bibitem{Chiaretti2014}
Chiaretti, S., Zini, G., Bassan, R.: Diagnosis and subclassification of acute
  lymphoblastic leukemia. Mediterranean journal of hematology and infectious
  diseases  \textbf{6}(1),  e2014073--e2014073 (Nov 2014),
  \url{https://www.ncbi.nlm.nih.gov/pubmed/25408859}

\bibitem{dataset-4}
Duggal, R., Gupta, A., Gupta, R.: Segmentation of overlapping/touching white
  blood cell nuclei using artificial neural networks. In: CME Series on
  Hemato-Oncopathology. All India Institute of Medical Sciences (AIIMS), New
  Delhi, India (2016)

\bibitem{dataset-5}
Duggal, R., Gupta, A., Gupta, R., Mallick, P.: Sd-layer: Stain deconvolutional
  layer for cnns in medical microscopic imaging. In: Descoteaux, M.,
  Maier-Hein, L., Franz, A., Jannin, P., Collins, D.L., Duchesne, S. (eds.)
  Medical Image Computing and Computer Assisted Intervention (MICCAI 2017). pp.
  435--443. Springer International Publishing, Cham (2017)

\bibitem{dataset-3}
Duggal, R., Gupta, A., Gupta, R., Wadhwa, M., Ahuja, C.: Overlapping cell
  nuclei segmentation in microscopic images using deep belief networks. In:
  Proceedings of the Tenth Indian Conference on Computer Vision, Graphics and
  Image Processing. pp. 82:1--82:8. ICVGIP '16, ACM, New York, NY, USA (2016).
  \doi{10.1145/3009977.3010043},
  \url{http://doi.acm.org/10.1145/3009977.3010043}

\bibitem{dataset-1}
Gupta, A., Duggal, R., Gupta, R., Kumar, L., Thakkar, N., Satpathy, D.:
  {GCTI-SN}: Geometry-inspired chemical and tissue invariant stain
  normalization of microscopic medical images. Under review

\bibitem{dataset-2}
Gupta, R., Mallick, P., Duggal, R., Gupta, A., Sharma, O.: Stain color
  normalization and segmentation of plasma cells in microscopic images as a
  prelude to development of computer assisted automated disease diagnostic tool
  in multiple myeloma. Clinical Lymphoma, Myeloma and Leukemia  \textbf{17}(1),
  ~e99 (2019/03/18 2017)

\bibitem{Hu_2018_CVPR}
Hu, J., Shen, L., Sun, G.: Squeeze-and-excitation networks. In: The IEEE
  Conference on Computer Vision and Pattern Recognition (CVPR) (June 2018)

\bibitem{adam}
{Kingma}, D.P., {Ba}, J.: {Adam: A Method for Stochastic Optimization}. The
  International Conference on Learning Representations (ICLR '15)  (Dec 2015)

\bibitem{ALL-IDB2}
{Labati}, R.D., {Piuri}, V., {Scotti}, F.: All-idb: The acute lymphoblastic
  leukemia image database for image processing. In: 2011 18th IEEE
  International Conference on Image Processing. pp. 2045--2048 (Sep 2011).
  \doi{10.1109/ICIP.2011.6115881}

\bibitem{Madhloom2012}
{Madhloom}, H.T., {Kareem}, S.A., {Ariffin}, H.: A robust feature extraction
  and selection method for the recognition of lymphocytes versus acute
  lymphoblastic leukemia. In: 2012 International Conference on Advanced
  Computer Science Applications and Technologies (ACSAT). pp. 330--335 (Nov
  2012). \doi{10.1109/ACSAT.2012.62}

\bibitem{Mohapatra2014}
Mohapatra, S., Patra, D., Satpathy, S.: An ensemble classifier system for early
  diagnosis of acute lymphoblastic leukemia in blood microscopic images. Neural
  Computing and Applications  \textbf{24}(7),  1887--1904 (Jun 2014).
  \doi{10.1007/s00521-013-1438-3}

\bibitem{Pui2017}
Pui, C.H.: Acute Lymphoblastic Leukemia, pp. 39--43. Springer Berlin
  Heidelberg, Berlin, Heidelberg (2017). \doi{10.1007/978-3-662-46875-3\_57}

\bibitem{Putzu2013}
Putzu, L., Ruberto, C.D.: White blood cells identication and classication from
  leukemic blood image. In: International Work-Conference on Bioinformatics and
  Biomedical Engineering, {IWBBIO} 2013, Granada, Spain, March 18-20, 2013.
  Proceedings. pp. 99--106 (2013)

\bibitem{Rehman2018}
Rehman, A., Abbas, N., Saba, T., Rahman, S.I.u., Mehmood, Z., Kolivand, H.:
  Classification of acute lymphoblastic leukemia using deep learning.
  Microscopy Research and Technique  \textbf{81}(11),  1310--1317 (2018).
  \doi{10.1002/jemt.23139}

\bibitem{ILSVRC15}
Russakovsky, O., Deng, J., Su, H., Krause, J., Satheesh, S., Ma, S., Huang, Z.,
  Karpathy, A., Khosla, A., Bernstein, M., Berg, A.C., Fei-Fei, L.: {ImageNet
  Large Scale Visual Recognition Challenge}. International Journal of Computer
  Vision (IJCV)  \textbf{115}(3),  211--252 (2015).
  \doi{10.1007/s11263-015-0816-y}

\bibitem{Shafique2018}
Shafique, S., Tehsin, S.: Acute lymphoblastic leukemia detection and
  classification of its subtypes using pretrained deep convolutional neural
  networks. Technology in Cancer Research \& Treatment  \textbf{17},
  1533033818802789 (2018). \doi{10.1177/1533033818802789}, pMID: 30261827

\bibitem{Vogado2017}
{Vogado}, L.H.S., {Veras}, R.D.M.S., {Andrade}, A.R., {Araujo}, F.H.D.D.,
  e.~{Silva}, R.R.V., {Aires}, K.R.T.: Diagnosing leukemia in blood smear
  images using an ensemble of classifiers and pre-trained convolutional neural
  networks. In: 2017 30th SIBGRAPI Conference on Graphics, Patterns and Images
  (SIBGRAPI). pp. 367--373 (Oct 2017). \doi{10.1109/SIBGRAPI.2017.55}

\bibitem{Xie_2017_CVPR}
Xie, S., Girshick, R., Dollar, P., Tu, Z., He, K.: Aggregated residual
  transformations for deep neural networks. In: The IEEE Conference on Computer
  Vision and Pattern Recognition (CVPR) (July 2017)

\bibitem{Yosinski:2014:TFD:2969033.2969197}
Yosinski, J., Clune, J., Bengio, Y., Lipson, H.: How transferable are features
  in deep neural networks? In: Proceedings of the 27th International Conference
  on Neural Information Processing Systems - Volume 2. pp. 3320--3328. NIPS'14,
  MIT Press, Cambridge, MA, USA (2014),
  \url{http://dl.acm.org/citation.cfm?id=2969033.2969197}

\end{thebibliography}

\end{document}